\documentclass[12pt]{article} 
\usepackage{amsfonts,amsmath,amssymb,amsthm,dsfont,epsfig,graphicx,booktabs,latexsym} %
\usepackage{natbib}
\usepackage{enumerate}
\usepackage{multirow}
\usepackage{color}
\usepackage{amssymb,mathrsfs}
\usepackage{amsbsy,amsthm}
\usepackage{float,epsfig,subfigure}
\usepackage{multirow,setspace}
\usepackage{authblk,epstopdf}
\usepackage{array} 
\usepackage{paralist}
\usepackage{verbatim} 
\usepackage{arydshln}
\usepackage{algorithm,algpseudocode}
\algnewcommand{\And}{\textbf{and}}

\usepackage[margin=1in]{geometry}

\usepackage{makecell}

\begin{document}
\title{Simultaneous Long-tailed Recognition and Multi-modal Fusion for Highly Imbalanced Multi-modal Data}
\date{}

\author[a]{Heegeon Yoon}
\author[a]{Heeyoung Kim}
\affil[a]{\small{Department of Industrial and Systems Engineering, Korea Advanced Institute of Science and Technology (KAIST), 291 Daehak-ro, Yuseong-gu, Daejeon 34141, Republic of Korea}}
\affil[+]{Corresponding author, Email: heeyoungkim@kaist.ac.kr }
\maketitle

\begin{abstract}
   Long-tailed distributions in class-imbalanced data present a fundamental challenge for deep learning models, which tend to be biased toward majority classes. While recent methods for long-tailed recognition have mitigated this issue, they are largely restricted to single-modal inputs and cannot fully exploit complementary information from diverse data sources. In this work, we introduce a new framework for long-tailed recognition that explicitly handles multi-modal inputs. Our approach extends multi-expert architectures to the multi-modal setting by fusing heterogeneous data into a unified representation while leveraging modality-specific networks to estimate the informativeness of each modality. These confidence-guided weights dynamically modulate the fusion process, ensuring that more informative modalities contribute more strongly to the final decision. To further enhance performance, we design specialized training and test procedures that accommodate diverse modality combinations, including images and tabular data. Extensive experiments on benchmark and real-world datasets demonstrate that the proposed approach not only effectively integrates multi-modal information but also outperforms existing methods in handling long-tailed, class-imbalanced scenarios, highlighting its robustness and generalization capability.
\end{abstract}

\textbf{keyword}
long-tailed recognition, multi-modal fusion, multiple experts, self-supervised learning.

\section{Introduction} \label{s:intro}
Deep neural networks (DNNs) have achieved notable success in a wide range of tasks due to their strong representational and computational capabilities \citep{kim2021locally,kim2023contextual,koo2024deep,yoon2024uncertainty}. As datasets continue to expand in size and complexity, these models have become increasingly sophisticated, with deeper architectures and greater expressive power. Despite these advances, DNNs trained on imbalanced class distributions often exhibit a tendency to favor majority classes, leading to degraded performance on underrepresented classes \citep{lee2021abc, zhang2023deep,park2024rebalancing, lee2025learnable}. Because many real-world datasets follow long-tailed distributions in which minority classes can contain critical and informative patterns, developing methods that enable DNNs to learn effectively from imbalanced data is essential to prevent the loss of valuable information from these rare classes \citep{park2022prediction, yoon2023label, lee2023semi}.

Moreover, data encountered in real-world applications are frequently multi-modal, meaning that observations originate from heterogeneous sources \citep{choy2016looking,soh2018application,chung2019crime,yoon2026multimodal}. To make effective use of such heterogeneous inputs, a wide range of multi-modal learning approaches have been proposed that exploit complementary information across modalities to enhance predictive performance \citep{han2022multimodal,cho2023prediction}. Common strategies integrate multiple modalities into a unified representation, using techniques that span from straightforward feature-level concatenation \citep{lee2013dependence,hong2020more, huang2021makes} to more sophisticated neural architectures that learn joint representations in an end-to-end manner \citep{li2021multi, wang2021mogonet}.

Although prior research has extensively studied class imbalance and multi-modal data separately, relatively little attention has been given to settings where both challenges arise simultaneously. Developing methods that can effectively handle long-tailed class distributions in conjunction with multi-modal inputs is therefore essential in many real-world applications. In the medical domain, for instance, datasets often contain far more samples from healthy individuals than from patients with specific conditions, while also encompassing diverse data types such as imaging data (e.g., X-rays) alongside auxiliary information including demographics and clinical histories.

In this work, we propose a new framework for long-tailed recognition explicitly designed to handle multi-modal data. Unlike prior methods that address either class imbalance or multi-modal fusion independently, the proposed method jointly models both challenges through a principled integration of multi-expert learning and modality-aware aggregation. For the multi-expert backbone, the proposed method builds on the self-supervised aggregation of diverse experts (SADE) \citep{zhang2022self}, which is well-suited to handle varying test-time class distributions, including those that differ from the training distribution. SADE trains multiple expert networks, each specialized for a distinct class distribution, and adaptively combines their predictions at inference using self-supervised aggregation weights. However, SADE was originally limited to single-modal inputs, leaving a gap for multi-modal scenarios.

To address this, we redesign the expert networks to integrate multi-modal inputs into unified latent representations. Beyond a naive fusion, we introduce an auxiliary modality-assessment module consisting of modality-specific networks that explicitly quantify the informativeness of each modality, inspired by multi-modal dynamics (MMD) \citep{han2022multimodal}. Each network outputs a confidence score reflecting the classification reliability of its modality, which is then used to weight the modality’s contribution during fusion. This weighting mechanism goes beyond conventional multi-modal fusion approaches, including MMD, as it is tightly integrated with the multi-expert framework and jointly optimized under long-tailed learning objectives. In this way, the proposed method not only aggregates complementary information across modalities but also dynamically emphasizes informative modalities in a manner that is conditioned on both the input and the class distribution.

Lastly, we design training and testing procedures specifically tailored to handle diverse combinations of modalities, ensuring effective multi-modal fusion. Unlike the original SADE, which relies on image data augmentations for self-supervised test-time adaptation, generating comparable augmentations for tabular or non-image modalities is often infeasible. To address this, we update the aggregation weights exclusively during training rather than at test time. Leveraging the availability of training labels, we optimize these weights in a supervised manner after training multiple expert networks, eliminating the need for data augmentations. The learned aggregation weights are then applied directly at test time. This approach relies on the assumption that the training and test datasets share the same distribution, as the aggregation weights are tuned to the long-tailed class distribution observed in the training data.

Our contributions can be summarized as follows:\\
(i) We propose a new long-tailed recognition framework that simultaneously handles multi-modal inputs and class imbalance. \\
(ii) We design specialized training and testing strategies to accommodate diverse combinations of input modalities. \\
(iii) We conduct extensive evaluations against multiple baseline methods, demonstrating the effectiveness of the proposed approach.

The remainder of this paper is organized as follows. Section~\ref{related} reviews prior work on handling multi-modal inputs and addressing class imbalance. 
Section~\ref{s:review} provides a brief overview of SADE and MMD. 
Section~\ref{s:method} describes the proposed model along with its training and testing procedures. Section~\ref{s:experiment} presents experimental results on both benchmark and real-world datasets. Finally, Section~\ref{s:conclusion} concludes the paper.

\section{Related Work}\label{related}
\textbf{Class-Imbalanced Learning.} Long-tailed recognition remains challenging because models trained on imbalanced datasets tend to overfit majority classes. Traditional solutions include resampling and reweighting strategies \citep{chawla2002smote, alani2020classifying, li2022imbalanced}, while recent multi-expert frameworks train specialized networks on different class distributions and combine them via adaptive weighting \citep{xiang2020learning, cai2021ace, cui2022reslt}. Notably, SADE \citep{zhang2022self} constructs three experts—long-tailed, uniform, and inversely long-tailed—and aggregates their predictions in a self-supervised manner at test time, achieving robustness across varying test-time distributions. However, these frameworks are restricted to single-modal inputs, limiting their applicability to modern multi-source datasets.

\textbf{Multi-Modal Learning.} Multi-modal fusion integrates heterogeneous data into unified representations to exploit complementary information \citep{wajid2021multimodal}. Fusion approaches include early (data-level), late (decision-level), and intermediate (feature-level) strategies, with intermediate fusion being the most common due to its ability to capture both intra- and inter-modal interactions \citep{suresh2017clinical, zadeh2017tensor, han2022multimodal}. Recent methods, such as MMD \citep{han2022multimodal}, dynamically weigh modalities by estimating their informativeness per sample, mitigating the impact of noisy or uninformative inputs. While effective, these approaches rarely consider class imbalance and are not designed to leverage distribution-specialized experts.

Existing methods either address long-tailed distributions or multi-modal fusion, but few explicitly handle both simultaneously. Multi-expert models like SADE do not support multi-modal inputs, and multi-modal fusion methods like MMD are agnostic to long-tailed distributions.

\section{Background} \label{s:review}

\subsection{\emph{SADE}} \label{s:imbalance}
SADE \citep{zhang2022self} is a robust framework for class-imbalanced learning that trains three skill-diverse experts, $E_1$, $E_2$, and $E_3$, each specializing in a different class distribution: long-tailed, uniform, and inversely long-tailed, respectively. 
To induce these distributional preferences, the experts are trained with distinct loss functions. 
Specifically, $E_1$ uses the standard softmax cross-entropy loss $\mathcal{L}_{ce}$, $E_2$ employs the balanced softmax loss $\mathcal{L}_{bal}$ \citep{ren2020balanced}, and $E_3$ is trained with the inverse softmax loss $\mathcal{L}_{inv}$:
$\mathcal{L}_{ce}= \frac{1}{n_s} \sum_{x_i \in \mathcal{D}_s} -y_i \log \sigma(v_1(x_i))$, 
$\mathcal{L}_{bal} = \frac{1}{n_s} \sum_{x_i \in \mathcal{D}_s} -y_i \log \sigma(v_2(x_i)+\log \pi)$, and $\mathcal{L}_{inv} = \frac{1}{n_s} \sum_{x_i \in \mathcal{D}_s} -y_i \log \sigma(v_3(x_i)+\log \pi-\log \bar{\pi})$,
where $\mathcal{D}_s=\{(x_i,y_i)\}_{i=1}^{n_s}$ denotes the long-tailed training set, $\sigma(\cdot)$ is the softmax function, $v_j(\cdot)$ is the logit output of expert $E_j$, $\pi$ is the label frequency vector, and $\bar{\pi}$ is its reversed version. 
The experts are trained jointly using the composite loss $\mathcal{L}=\mathcal{L}_{ce}+\mathcal{L}_{bal}+\mathcal{L}_{inv}$.

At test time, SADE adaptively aggregates the experts by learning mixture weights in a self-supervised manner, as class labels are unavailable. 
Given an input $x$, two stochastic augmentations $x^1$ and $x^2$ are generated. 
For each $x^k$, the aggregated prediction is computed as
$\hat{y}^k = \sum_{j=1}^3 w_j v_j(x^k)$ with $\sum_{j=1}^3 w_j = 1$, 
where $\{w_j\}$ are trainable aggregation weights. 
The weights are optimized by maximizing prediction stability between the two augmented views:
\begin{equation}\label{stable}
\max_{w} \frac{1}{n_t} \sum_{x \in \mathcal{D}_t} \hat{y}^1 \cdot \hat{y}^2,
\end{equation}
where $n_t$ denotes the number of data points in the test set $\mathcal{D}_t$.

\subsection{\emph{MMD}} \label{s:fusion}

MMD \citep{han2022multimodal} is an intermediate fusion framework that emphasizes informative modalities by quantifying their sample-wise relevance.
Given a multi-modal dataset $\{\{\mathbf{x}_i^m\}_{m=1}^M,y_i\}_{i=1}^N$, MMD assumes a modality-specific classifier $f^m:\mathbf{x}_i^m \mapsto y_i$, where $\mathbf{x}_i^m \in \mathbb{R}^{d_m}$ and $y_i \in \{1,\dots,K\}$.
The informativeness of modality $m$ for sample $i$ is measured by its True Class Probability (TCP),
\begin{equation}\label{tcp}
TCP_i^m=\mathbf{p}^m(y=y_i|\mathbf{x}_i^m),
\end{equation}
where $\mathbf{p}^m(y|\mathbf{x}_i^m)=[p^m(y=1|\mathbf{x}_i^m),\ldots,p^m(y=K|\mathbf{x}_i^m)]$ represents a categorical distribution obtained from the softmax output of $f^m$.
Since TCP depends on ground-truth labels and is available only during training, MMD introduces a confidence network $g^m:\mathbf{x}_i^m \mapsto TCP_i^m$, sharing all layers except the output with $f^m$, to approximate TCP.
In the fusion layer, the extracted representation $\mathbf{h}_i^m=f_1^m(\mathbf{x}_i^m)$, where $f_1^m$ denotes the shared sub-network, is weighted by $\widehat{TCP_i^m}=g^m(\mathbf{x}_i^m)$, an approximation of $TCP_i^m$, to produce the final concatenated representation $\mathbf{h}_i=[\widehat{TCP_i^1}\mathbf{h}_i^1,...,\widehat{TCP_i^M}\mathbf{h}_i^M]$. This fusion layer is followed by a classifier $f:\mathbf{h}_i \xrightarrow{} y_i$ for the final prediction.

To train the model, MMD employs a unified loss function defined as follows:
\begin{equation}\label{eq.6}
\mathcal{L}=\mathcal{L}^f+\lambda \sum_{m=1}^M (\mathcal{L}^{f^m}+\mathcal{L}^{g^m}),
\end{equation}
where $\mathcal{L}^f=-\frac{1}{N}\sum_{i=1}^{N}\mathcal{L}_{ce}(y_i,f(\mathbf{h}_i))$ and $\mathcal{L}^{f^m}=-\frac{1}{N}\sum_{i=1}^{N}\mathcal{L}_{ce}(y_i,f^m(\mathbf{x}_i^m))$ are the cross-entropy losses for the classifiers $f$ and $f^m$, respectively, $\mathcal{L}^{g^m}=\frac{1}{N}\sum_{i=1}^{N}(\widehat{TCP_i^m}-TCP_i^m)^2$ is the $\ell_2$ loss for approximating TCP, and $\lambda$ is a hyperparameter that balances the loss components.

\section{Methodology} \label{s:method}
We propose a unified framework that combines distribution-specialized experts with modality-aware fusion. The proposed method integrates multi-modal inputs within each expert and adaptively weights modalities during fusion, enabling robust recognition on datasets that are both multi-modal and class-imbalanced.

\subsection{\emph{Problem scenario}} \label{s:problem}
Our training dataset consists of multi-modal data points, which are represented as pairs  \\ $\mathcal{D}_{tr}=\{(\{\mathbf{x}_1^1,...,\mathbf{x}_1^M\},y_1),...,(\{\mathbf{x}_N^1,...,\mathbf{x}_N^M\},y_N)\}$, where $\mathbf{x}_i^m\in\mathbb{R}^{d_m}$ is the $i$th instance of the $m$th modality, and $y_i\in\{1,...,K\}$ is the corresponding label. The dataset contains $N$ samples in total, with $N_k$ samples in class $k$, such that $N=\sum_{k=1}^K N_k$. We assume that the dataset is class-imbalanced with a long-tailed distribution, wherein classes vary in size; specifically, if sorted by class size in descending order, i.e., $N_i\geq N_j$ for $i<j$, then $N_1\gg N_K$. The imbalance ratio is defined as $r=N_1/N_K$. The test dataset $\mathcal{D}_{te}$ is structured similarly. The objective of our model is to predict the class label $y$ of an unknown instance $\{\mathbf{x}^1,...,\mathbf{x}^M\}$ in the test dataset $\mathcal{D}_{te}$ using the long-tailed training dataset $\mathcal{D}_{tr}$. For simplicity, we illustrate our method using the case of two modalities ($M=2$).

\subsection{\emph{Modality-aware multi-expert learning}} \label{s:proposed}
We propose a framework for long-tailed recognition that enables simultaneous multi-modal fusion and effective handling of highly imbalanced multi-modal data. Building on the structure of SADE \citep{zhang2022self}, our approach constructs three independent experts, each specializing in learning from long-tailed, uniform, or inversely long-tailed distributions. Since SADE was originally designed for single-modal data, we adapt it by modifying the network architecture of each expert to accept multi-modal inputs, thereby enabling multi-modal fusion within the SADE framework. This fusion results in a unified single-modal representation.

\begin{figure}[htbp]
	\vskip 0.2in
	\begin{center}
		\centerline{\includegraphics[width=5.0in]{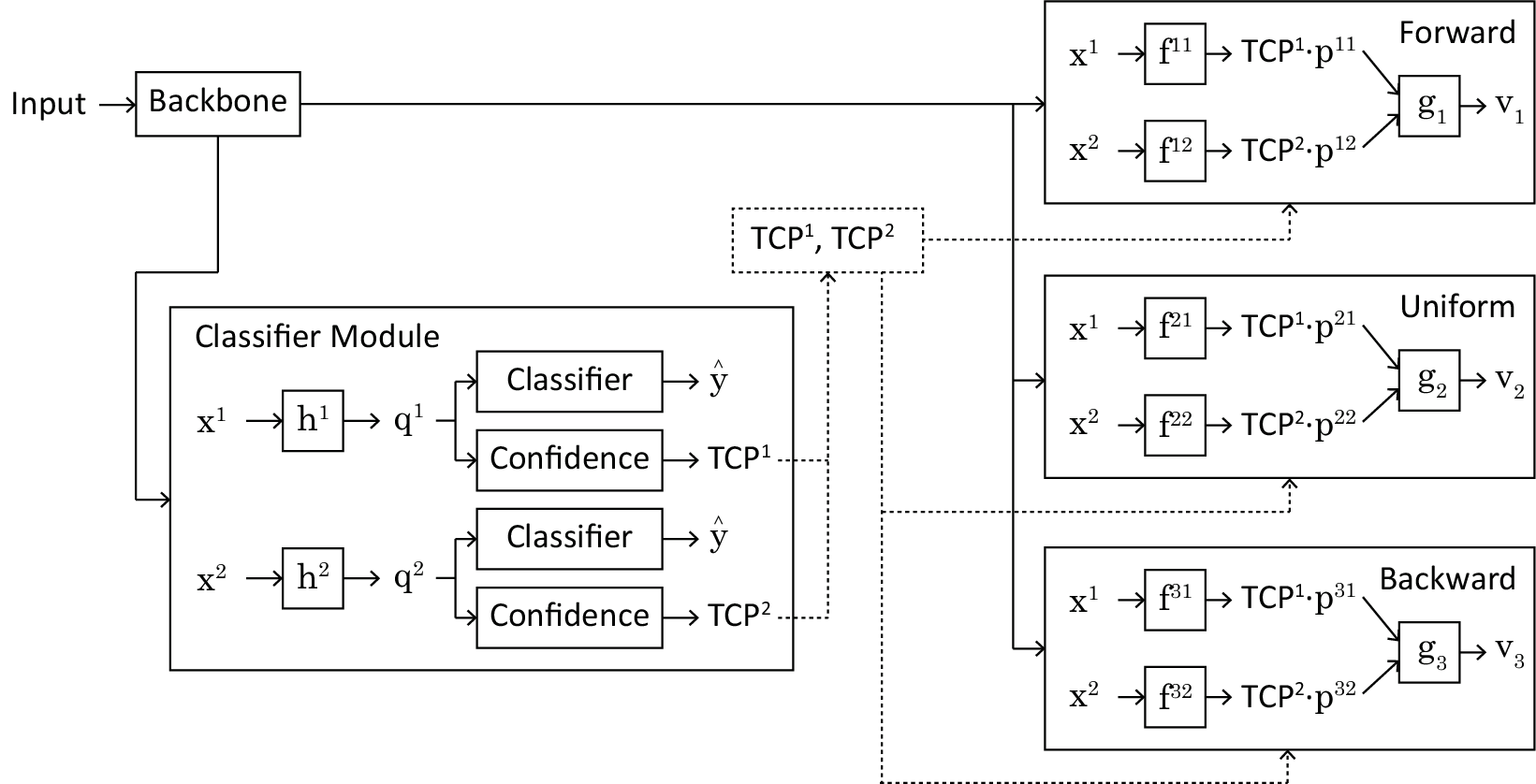}}
		\caption{Structure of the proposed model with the classifier module, illustrated with two input modalities.}
		\label{fig:proposed}
	\end{center}
	\vskip -0.2in
\end{figure}

To further enhance multi-modal fusion, we quantify the informativeness of each modality using the TCP defined in Eq.~\eqref{tcp}. 
To implement TCP-based weighting, we introduce a classifier module that computes the TCP for each modality. This module consists of modality-specific networks, each comprising two sub-networks that share all layers except the output layer. Given an input $\mathbf{x}^m$ ($m=1,2$), each sub-network performs forward propagation and outputs either a predicted label $\hat{y}$ or the corresponding TCP. The TCP derived from the sub-network predicting $\hat{y}$ is then used by each expert to weight each modality prior to fusion. The overall architecture of the proposed model, including the classifier module, is illustrated in Figure~\ref{fig:proposed}.

\subsection{\emph{Training and test procedures}} \label{s:cases}
\subsubsection{\emph{Case 1: Fully image-based modalities}} \label{s:case1}
\begin{figure}[ht]
	\vskip 0.2in
	\begin{center}
		\centerline{\includegraphics[width=5.0in]{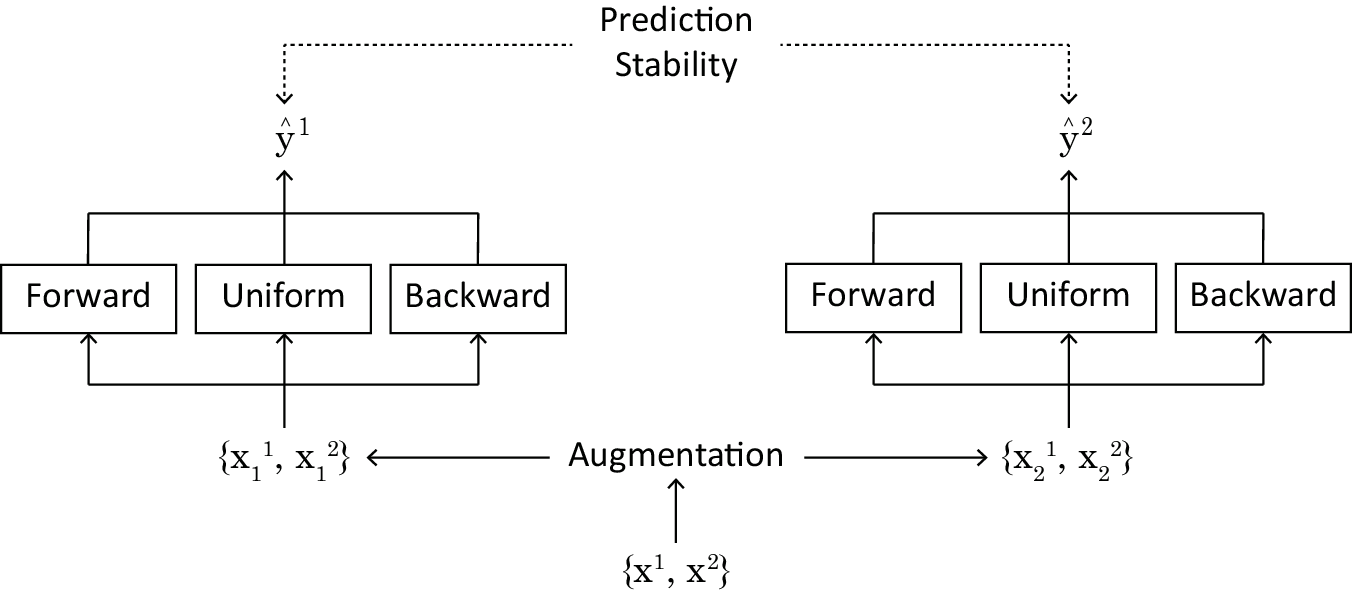}}
		\caption{Test-time training procedure of the proposed model.}
		\label{fig:test1}
	\end{center}
	\vskip -0.2in
\end{figure}

After training the three experts, we adopt a test-time training strategy inspired by SADE \citep{zhang2022self} to optimize the aggregation weights $\{w_1,w_2,w_3\}$ of the experts. This test-time training procedure, illustrated in Figure~\ref{fig:test1}, consists of three main steps:

1) Data augmentation---For a given multi-modal sample $\{\mathbf{x}^1,\mathbf{x}^2\}$, two probabilistic data augmentations $\{\mathbf{x}_1^1,\mathbf{x}_1^2\}$ and $\{\mathbf{x}_2^1,\mathbf{x}_2^2\}$ are generated.

2) Expert aggregation---For each augmented sample $\{\mathbf{x}_i^1,\mathbf{x}_i^2\}$, the final softmax prediction $\hat{y}^i$ is obtained as a weighted linear combination of the experts' output logits: 
$$\hat{y}^i=w_1\cdot v_1(\{\mathbf{x}_i^1,\mathbf{x}_i^2\})+w_2\cdot v_2(\{\mathbf{x}_i^1,\mathbf{x}_i^2\})+w_3\cdot v_3(\{\mathbf{x}_i^1,\mathbf{x}_i^2\}).$$

3) Prediction stability maximization---Given the predictions $\hat{y}^1$ and $\hat{y}^2$ for the two augmented samples, the loss function defined in Eq.\eqref{stable} is used to optimize the aggregation weights.

\subsubsection{\emph{Case 2: Mixed image and tabular modalities}} \label{s:case2}
The test-time training strategy described in Section~\ref{s:case1} is applicable only to image modalities, as it relies on stochastic data augmentations. While a wide range of probabilistic image augmentation techniques are well established in computer vision, comparable augmentation methods for non-image data types, such as tabular data, are less mature. Moreover, naive interpolation of tabular data may degrade model performance, particularly when samples lie near decision boundaries, where even small perturbations in input values can lead to substantially different predictions.

To address this limitation, we eliminate the need for data augmentation by learning the experts’ aggregation weights during the training phase. Since ground-truth labels are available for all training samples, prediction stability maximization is unnecessary; instead, we optimize the aggregation weights using a standard softmax cross-entropy loss. 
However, this approach assumes that the training and test datasets follow the same distribution. This assumption is reasonable in practice, as data are often collected from the same environment.

\begin{figure}[htbp]
	\vskip 0.2in
	\begin{center}
		\centerline{\includegraphics[width=4.0in]{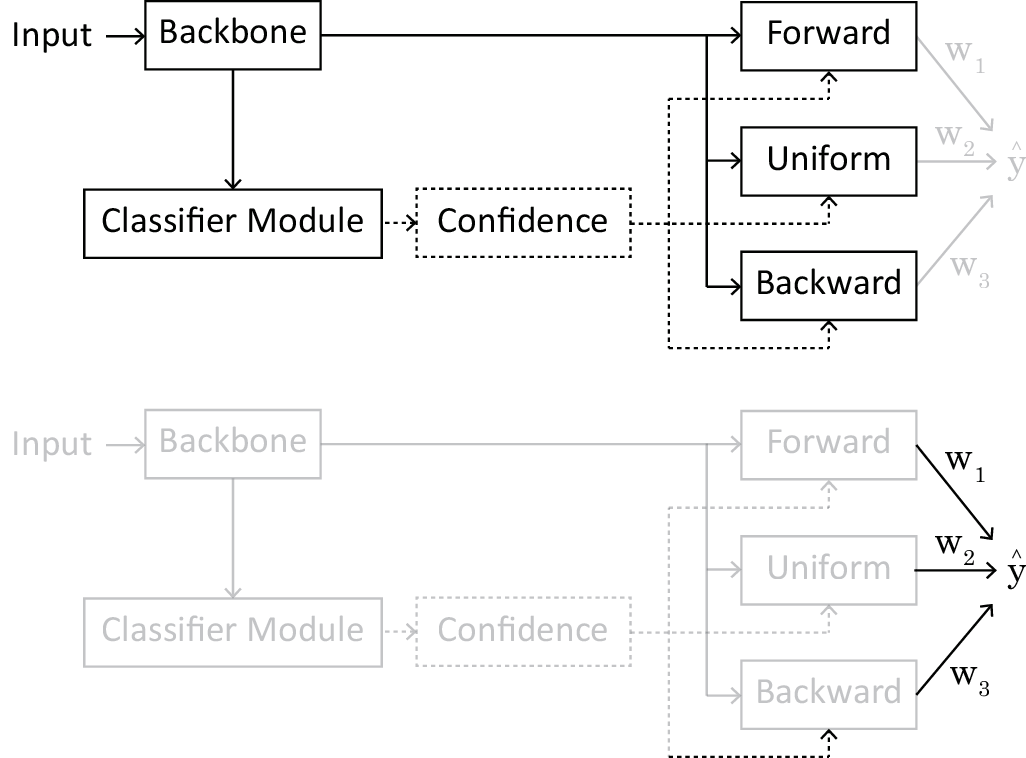}}
		\caption{Modified training strategy with one image modality and one tabular modality. The top (bottom) figure illustrates Phase 1 (Phase 2). In both figures, the training stages are highlighted in bold.}
		\label{fig:test2}
	\end{center}
	\vskip -0.2in
\end{figure}

Figure~\ref{fig:test2} illustrates the modified training process of the proposed model, which consists of two phases. In the first phase, the three experts are trained using the TCP computed by the classifier module, as shown in Figure~\ref{fig:proposed}. In the second phase, the aggregation weights of the trained experts are optimized using the training labels. With this modified training strategy, no optimization is performed during the test stage; instead, the test stage is used solely to evaluate the performance of the trained model, consistent with conventional training paradigms.

\section{Experiments} \label{s:experiment}
In this section, we evaluate the effectiveness of the proposed method through experiments on two groups of datasets. For the setting with two image modalities, we use a combination of two benchmark image datasets, \textit{MNIST} and \textit{SVHN}. For the setting with one image modality and one tabular modality, we use a real-world medical dataset from the SIIM–ISIC Melanoma Classification Challenge hosted on Kaggle. 

Model performance is evaluated using two metrics: accuracy and F1 score. Accuracy measures the proportion of correct predictions over the entire dataset and is computed as the ratio of correctly classified samples to the total number of samples. The F1 score, defined as the harmonic mean of precision and recall, balances these two quantities, where precision denotes the proportion of true positive predictions among all positive predictions, and recall denotes the proportion of true positive predictions among all actual positive samples. Unlike accuracy, which weights all classes equally, the F1 score places greater emphasis on minority classes and is therefore more suitable for evaluating long-tailed recognition performance. 

Experimental results on the benchmark image datasets and the real-world medical dataset are presented in Sections~\ref{s:benchmarks} and~\ref{s:melanoma}, respectively.

\subsection{\emph{Experiments on \textit{MNIST} and \textit{SVHN} datasets}} \label{s:benchmarks}
\subsubsection{\emph{Datasets and experimental setup}} \label{s:benchmarks_setup}
We first conducted a series of experiments on a mixture of the \textit{MNIST} and \textit{SVHN} datasets. The \textit{MNIST} dataset consists of 10 classes of hand-written digits and comprises 60,000 training images and 10,000 test images, each with dimensions of $28\times 28$. The \textit{SVHN} dataset contains 10 classes of house numbers and consists of 73,257 training images and 26,032 test images, each with dimensions of $32 \times 32$. The dataset for multi-modal fusion was created by selecting an equal number of images from each of the two datasets for every class label and combining them, resulting in one \textit{MNIST} image and one \textit{SVHN} image per label.

Additionally, because the combined dataset was balanced, we manually created its imbalanced version by following the process described in \cite{cao2019learning} for generating long-tailed imbalanced versions of the \textit{CIFAR-10} and \textit{CIFAR-100} datasets, where the sample sizes for different classes follow an exponential decay pattern. 
In this study, we varied the training imbalance ratio $r$ from 10 to 100 to assess the robustness of our algorithm. Additionally, we included the case when the dataset is balanced ($r=1$) to verify that the proposed model performs effectively in a balanced scenario as well.

\textbf{Baselines.} We compared the performance of the proposed model with the following approaches that simultaneously perform multi-modal fusion and long-tailed recognition: (i) M$^2$LC-Net \citep{ou2021m}---trains a typical intermediate multi-modal fusion model with a  combination of the class-balanced loss function \citep{cui2019class} and the focal loss function \citep{lin2017focal} in a multi-disease classification task; and (ii) MMD-LA \citep{cho2023prediction}---utilizes MMD \citep{han2022multimodal} with the logit-adjusted cross-entropy loss function \citep{menon2020long} to address class imbalance in the  semiconductor memory module manufacturing process. Additionally, we devised a simple model called (iii) SADE-LMF, which extends SADE \citep{zhang2022self} by incorporating the low-rank multi-modal fusion concept from LMF \citep{liu2018efficient}, and used it as another baseline for comparison. Furthermore, we compared the proposed model with the original (iv) MMD \citep{han2022multimodal} and the original (v) SADE \citep{zhang2022self} to verify the  robustness and advantages of our approach when leveraging multiple modalities. Since SADE is designed for single-modal datasets, we used only one image modality as input for SADE during its training. 

\textbf{Network structure and optimization.} In each expert within our model, features from  \textit{MNIST} and \textit{SVHN} images were extracted using two convolutional layers, each followed by a max-pooling layer. During training, the extracted features from the two modalities were multiplied by the TCPs computed from the classifier module and then  concatenated to form a feature vector. This vector was then fed into the final classifier. The confidence networks within the classifier module were modeled using a single linear layer, which outputted a number representing the TCP of the corresponding modality. The initial weight parameters were randomly sampled from $\mathcal{N}(\boldsymbol{0},0.001^2 \boldsymbol{I})$, and the initial bias parameters were set to 0. We optimized the objective function for the proposed model using Adam with a batch size of 128, decay parameters for the first and second moments of the gradients $(\beta_1, \beta_2) = (0.9, 0.999)$, and a learning rate of $10^{-4}$.

\subsubsection{\emph{Classification results on \textit{MNIST} and \textit{SVHN}}} \label{s:benchmarks_results}
In addition to varying the training data distributions, we evaluated the robustness of the proposed method under different test data distributions. Specifically, we varied the distribution of test data from a long-tailed distribution with an imbalance ratio of 50 to an inversely long-tailed distribution with an imbalance ratio 50. For each training and test data distribution, we repeated experiment 10 times. In each experiment, we randomly selected the required number of images for each class label without replacement to satisfy the conditions for the long-tailed imbalance. The classification accuracies and F1 scores with standard errors for five test data distributions are shown in Table \ref{table:accuracy_mnist_svhn}.
\begin{table}[htbp]
\caption{\footnotesize{Classification accuracies and F1 scores, along with standard errors, for the \textit{MNIST} and \textit{SVHN} datasets. The abbreviation ``IR" denotes the imbalance ratio of the training data. The distribution of the test data spans from a long-tailed distribution with an imbalance ratio 50 (Forward 50) to an inversely long-tailed distribution with an imbalance ratio 50 (Backward 50).}}
\label{table:accuracy_mnist_svhn}
\vskip 0.15in
\begin{center}
\begin{small}
\begin{sc}
    \resizebox{16cm}{!}{%
    \begin{tabular}{cccccccccccc}
        \hline
        \multirow{2}{*}{Model} & \multirow{2}{*}{IR} & \multicolumn{2}{c}{Forward 50} & \multicolumn{2}{c}{Forward 10} & \multicolumn{2}{c}{Uniform} & \multicolumn{2}{c}{Backward 10} & \multicolumn{2}{c}{Backward 50} \\ \cline{3-12} & & Acc & F1 & Acc & F1 & Acc & F1 & Acc & F1 & Acc & F1 \\ \hline
        \multirow{4}{*}{MMD} & 100 & \begin{tabular}[c]{@{}c@{}}99.16\\ (0.06)\end{tabular} & \begin{tabular}[c]{@{}c@{}}0.8723 \\(0.0339)\end{tabular} & \begin{tabular}[c]{@{}c@{}}99.00 \\(0.11)\end{tabular} & \begin{tabular}[c]{@{}c@{}}0.8285\\ (0.0281)\end{tabular} & \begin{tabular}[c]{@{}c@{}}97.67 \\(0.22)\end{tabular} & \begin{tabular}[c]{@{}c@{}}0.6507 \\(0.0246)\end{tabular} & \begin{tabular}[c]{@{}c@{}}95.57 \\(0.21)\end{tabular} & \begin{tabular}[c]{@{}c@{}}0.4903 \\(0.0214)\end{tabular} & \begin{tabular}[c]{@{}c@{}}95.02 \\(0.21)\end{tabular} & \begin{tabular}[c]{@{}c@{}}0.3196 \\(0.0341)\end{tabular} \\
        & 50 & \begin{tabular}[c]{@{}c@{}}99.34 \\(0.10)\end{tabular} & \begin{tabular}[c]{@{}c@{}}0.8622 \\(0.0219)\end{tabular} & \begin{tabular}[c]{@{}c@{}}99.27 \\(0.04)\end{tabular} & \begin{tabular}[c]{@{}c@{}}0.8140 \\(0.0083)\end{tabular} & \begin{tabular}[c]{@{}c@{}}98.70 \\(0.04)\end{tabular} & \begin{tabular}[c]{@{}c@{}}0.6914 \\(0.0099)\end{tabular} & \begin{tabular}[c]{@{}c@{}}97.76 \\(0.09)\end{tabular} & \begin{tabular}[c]{@{}c@{}}0.5053 \\(0.0300)\end{tabular} & \begin{tabular}[c]{@{}c@{}}97.64 \\(0.43)\end{tabular} & \begin{tabular}[c]{@{}c@{}}0.5199\\ (0.0275)\end{tabular} \\
        & 10 & \begin{tabular}[c]{@{}c@{}}99.86 \\(0.05)\end{tabular} & \begin{tabular}[c]{@{}c@{}}0.9436 \\(0.0263)\end{tabular} & \begin{tabular}[c]{@{}c@{}}99.73\\ (0.03)\end{tabular} & \begin{tabular}[c]{@{}c@{}}0.8817\\ (0.0166)\end{tabular} & \begin{tabular}[c]{@{}c@{}}99.33 \\(0.03)\end{tabular} & \begin{tabular}[c]{@{}c@{}}0.7521 \\(0.0073)\end{tabular} & \begin{tabular}[c]{@{}c@{}}99.17\\ (0.07)\end{tabular} & \begin{tabular}[c]{@{}c@{}}0.6923\\ (0.0218)\end{tabular} & \begin{tabular}[c]{@{}c@{}}98.92 \\(0.11)\end{tabular} & \begin{tabular}[c]{@{}c@{}}0.6183\\ (0.0326)\end{tabular} \\
        & 1 & \begin{tabular}[c]{@{}c@{}}99.79 \\(0.03)\end{tabular} & \begin{tabular}[c]{@{}c@{}}0.9044\\ (0.0163)\end{tabular} & \begin{tabular}[c]{@{}c@{}}99.81\\ (0.02)\end{tabular} & \begin{tabular}[c]{@{}c@{}}0.9108 \\(0.0071)\end{tabular} & \begin{tabular}[c]{@{}c@{}}99.71\\ (0.03)\end{tabular} & \begin{tabular}[c]{@{}c@{}}0.8538 \\(0.0217)\end{tabular} & \begin{tabular}[c]{@{}c@{}}99.61\\ (0.03)\end{tabular} & \begin{tabular}[c]{@{}c@{}}0.8220\\ (0.0142)\end{tabular} & \begin{tabular}[c]{@{}c@{}}99.70 \\(0.03)\end{tabular} & \begin{tabular}[c]{@{}c@{}}0.8840 \\(0.0076)\end{tabular} \\ \hline
        \multirow{4}{*}{SADE} & 100 & \begin{tabular}[c]{@{}c@{}}91.82 \\(3.86)\end{tabular} & \begin{tabular}[c]{@{}c@{}}0.6916 \\(0.0007)\end{tabular} & \begin{tabular}[c]{@{}c@{}}90.98 \\(7.19)\end{tabular} & \begin{tabular}[c]{@{}c@{}}0.8725\\ (0.0012)\end{tabular} & \begin{tabular}[c]{@{}c@{}}88.96 \\(3.89)\end{tabular} & \begin{tabular}[c]{@{}c@{}}0.8886 \\(0.0004)\end{tabular} & \begin{tabular}[c]{@{}c@{}}88.43 \\(6.94)\end{tabular} & \begin{tabular}[c]{@{}c@{}}0.8704 \\(0.0009)\end{tabular} & \begin{tabular}[c]{@{}c@{}}88.06 \\(5.04)\end{tabular} & \begin{tabular}[c]{@{}c@{}}0.8019 \\(0.0012)\end{tabular} \\
        & 50 & \begin{tabular}[c]{@{}c@{}}92.45\\ (2.23)\end{tabular} & \begin{tabular}[c]{@{}c@{}}0.7004 \\(0.0006)\end{tabular} & \begin{tabular}[c]{@{}c@{}}89.52\\ (5.92)\end{tabular} & \begin{tabular}[c]{@{}c@{}}0.8490 \\(0.0010)\end{tabular} & \begin{tabular}[c]{@{}c@{}}85.37 \\(9.38)\end{tabular} & \begin{tabular}[c]{@{}c@{}}0.8513 \\(0.0011)\end{tabular} & \begin{tabular}[c]{@{}c@{}}82.99 \\(6.08)\end{tabular} & \begin{tabular}[c]{@{}c@{}}0.8224 \\(0.0008)\end{tabular} & \begin{tabular}[c]{@{}c@{}}84.41 \\(2.15)\end{tabular} & \begin{tabular}[c]{@{}c@{}}0.7804 \\(0.0004)\end{tabular} \\
        & 10 & \begin{tabular}[c]{@{}c@{}}92.67\\ (6.87)\end{tabular} & \begin{tabular}[c]{@{}c@{}}0.7007 \\(0.0006)\end{tabular} & \begin{tabular}[c]{@{}c@{}}91.38\\ (2.24)\end{tabular} & \begin{tabular}[c]{@{}c@{}}0.8776 \\(0.0003)\end{tabular} & \begin{tabular}[c]{@{}c@{}}89.22 \\(3.36)\end{tabular} & \begin{tabular}[c]{@{}c@{}}0.8915\\ (0.0004)\end{tabular} & \begin{tabular}[c]{@{}c@{}}87.95 \\(7.62)\end{tabular} & \begin{tabular}[c]{@{}c@{}}0.8607 \\(0.0010)\end{tabular} & \begin{tabular}[c]{@{}c@{}}87.80 \\(2.01)\end{tabular} & \begin{tabular}[c]{@{}c@{}}0.7968 \\(0.0007)\end{tabular} \\
        & 1 & \begin{tabular}[c]{@{}c@{}}93.12\\ (5.69)\end{tabular} & \begin{tabular}[c]{@{}c@{}}0.7218 \\(0.0004)\end{tabular} & \begin{tabular}[c]{@{}c@{}}92.64 \\(4.65)\end{tabular} & \begin{tabular}[c]{@{}c@{}}0.8945 \\(0.0010)\end{tabular} & \begin{tabular}[c]{@{}c@{}}92.48 \\(2.89)\end{tabular} & \begin{tabular}[c]{@{}c@{}}0.9246 \\(0.0003)\end{tabular} & \begin{tabular}[c]{@{}c@{}}92.64 \\(1.62)\end{tabular} & \begin{tabular}[c]{@{}c@{}}0.9046 \\(0.0002)\end{tabular} & \begin{tabular}[c]{@{}c@{}}92.97 \\(3.62)\end{tabular} & \begin{tabular}[c]{@{}c@{}}0.8538 \\(0.0006)\end{tabular} \\ \hline
        \multirow{4}{*}{MMD-LA} & 100 & \begin{tabular}[c]{@{}c@{}}99.31 \\(0.07)\end{tabular} & \begin{tabular}[c]{@{}c@{}}0.8426\\ (0.0383)\end{tabular} & \begin{tabular}[c]{@{}c@{}}99.25 \\(0.08)\end{tabular} & \begin{tabular}[c]{@{}c@{}}0.7499\\ (0.0316)\end{tabular} & \begin{tabular}[c]{@{}c@{}}98.19 \\(0.09)\end{tabular} & \begin{tabular}[c]{@{}c@{}}0.6552 \\(0.0247)\end{tabular} & \begin{tabular}[c]{@{}c@{}}97.37\\ (0.20)\end{tabular} & \begin{tabular}[c]{@{}c@{}}0.4251\\ (0.0244)\end{tabular} & \begin{tabular}[c]{@{}c@{}}96.65 \\(0.35)\end{tabular} & \begin{tabular}[c]{@{}c@{}}0.4017 \\(0.0309)\end{tabular} \\
        & 50 & \begin{tabular}[c]{@{}c@{}}99.63\\ (0.05)\end{tabular} & \begin{tabular}[c]{@{}c@{}}0.8925 \\(0.0489)\end{tabular} & \begin{tabular}[c]{@{}c@{}}99.37\\ (0.05)\end{tabular} & \begin{tabular}[c]{@{}c@{}}0.8107\\ (0.0195)\end{tabular} & \begin{tabular}[c]{@{}c@{}}99.05 \\(0.07)\end{tabular} & \begin{tabular}[c]{@{}c@{}}0.6957 \\(0.0124)\end{tabular} & \begin{tabular}[c]{@{}c@{}}98.63\\ (0.14)\end{tabular} & \begin{tabular}[c]{@{}c@{}}0.5537\\ (0.0245)\end{tabular} & \begin{tabular}[c]{@{}c@{}}98.22 \\(0.19)\end{tabular} & \begin{tabular}[c]{@{}c@{}}0.5401\\ (0.0243)\end{tabular} \\
        & 10 & \begin{tabular}[c]{@{}c@{}}99.82 \\(0.05)\end{tabular} & \begin{tabular}[c]{@{}c@{}}0.9162 \\(0.0223)\end{tabular} & \begin{tabular}[c]{@{}c@{}}99.73\\ (0.03)\end{tabular} & \begin{tabular}[c]{@{}c@{}}0.8768 \\(0.0149)\end{tabular} & \begin{tabular}[c]{@{}c@{}}99.50\\ (0.04)\end{tabular} & \begin{tabular}[c]{@{}c@{}}0.8004 \\(0.0178)\end{tabular} & \begin{tabular}[c]{@{}c@{}}99.37 \\(0.09)\end{tabular} & \begin{tabular}[c]{@{}c@{}}0.7617\\ (0.0322)\end{tabular} & \begin{tabular}[c]{@{}c@{}}99.19 \\(0.08)\end{tabular} & \begin{tabular}[c]{@{}c@{}}0.7288\\ (0.0344)\end{tabular} \\
        & 1 & \begin{tabular}[c]{@{}c@{}}99.89 \\(0.02)\end{tabular} & \begin{tabular}[c]{@{}c@{}}0.9589\\ (0.0167)\end{tabular} & \begin{tabular}[c]{@{}c@{}}99.81\\ (0.02)\end{tabular} & \begin{tabular}[c]{@{}c@{}}0.9084 \\(0.0088)\end{tabular} & \begin{tabular}[c]{@{}c@{}}99.73 \\(0.01)\end{tabular} & \begin{tabular}[c]{@{}c@{}}0.8600 \\(0.0110)\end{tabular} & \begin{tabular}[c]{@{}c@{}}99.76 \\(0.06)\end{tabular} & \begin{tabular}[c]{@{}c@{}}0.8848 \\(0.0389)\end{tabular} & \begin{tabular}[c]{@{}c@{}}99.79 \\(0.05)\end{tabular} & \begin{tabular}[c]{@{}c@{}}0.8859 \\(0.0295)\end{tabular} \\ \hline
        \multirow{4}{*}{SADE-LMF} & 100 & \begin{tabular}[c]{@{}c@{}}99.63 \\(1.96)\end{tabular} & \begin{tabular}[c]{@{}c@{}}0.7921\\ (0.0001)\end{tabular} & \begin{tabular}[c]{@{}c@{}}99.34 \\(1.47)\end{tabular} & \begin{tabular}[c]{@{}c@{}}0.9882\\ (0.0002)\end{tabular} & \begin{tabular}[c]{@{}c@{}}98.61\\ (0.81)\end{tabular} & \begin{tabular}[c]{@{}c@{}}0.9860 \\(0.0001)\end{tabular} & \begin{tabular}[c]{@{}c@{}}97.98 \\(1.20)\end{tabular} & \begin{tabular}[c]{@{}c@{}}0.9786\\ (0.0004)\end{tabular} & \begin{tabular}[c]{@{}c@{}}97.69 \\(1.20)\end{tabular} & \begin{tabular}[c]{@{}c@{}}0.9604 \\(0.0007)\end{tabular} \\
        & 50 & \begin{tabular}[c]{@{}c@{}}99.59 \\(0.90)\end{tabular} & \begin{tabular}[c]{@{}c@{}}0.8798\\ (0.0004)\end{tabular} & \begin{tabular}[c]{@{}c@{}}98.94 \\(2.65)\end{tabular} & \begin{tabular}[c]{@{}c@{}}0.9802 \\(0.0008)\end{tabular} & \begin{tabular}[c]{@{}c@{}}98.13\\ (1.73)\end{tabular} & \begin{tabular}[c]{@{}c@{}}0.9811 \\(0.0002)\end{tabular} & \begin{tabular}[c]{@{}c@{}}97.53 \\(4.59)\end{tabular} & \begin{tabular}[c]{@{}c@{}}0.9681 \\(0.0006)\end{tabular} & \begin{tabular}[c]{@{}c@{}}97.14 \\(6.81)\end{tabular} & \begin{tabular}[c]{@{}c@{}}0.9308\\ (0.0008)\end{tabular} \\ 
        & 10 & \begin{tabular}[c]{@{}c@{}}99.37 \\(1.83)\end{tabular} & \begin{tabular}[c]{@{}c@{}}0.7894 \\(0.0001)\end{tabular} & \begin{tabular}[c]{@{}c@{}}99.04\\ (0.90)\end{tabular} & \begin{tabular}[c]{@{}c@{}}0.9796 \\(0.0002)\end{tabular} & \begin{tabular}[c]{@{}c@{}}98.76\\ (1.17)\end{tabular} & \begin{tabular}[c]{@{}c@{}}0.9875 \\(0.0001)\end{tabular} & \begin{tabular}[c]{@{}c@{}}98.26\\ (1.30)\end{tabular} & \begin{tabular}[c]{@{}c@{}}0.9799 \\(0.0001)\end{tabular} & \begin{tabular}[c]{@{}c@{}}97.99 \\(2.69)\end{tabular} & \begin{tabular}[c]{@{}c@{}}0.9601 \\(0.0012)\end{tabular} \\
        & 1 & \begin{tabular}[c]{@{}c@{}}99.70 \\(1.47)\end{tabular} & \begin{tabular}[c]{@{}c@{}}0.8816 \\(0.0006)\end{tabular} & \begin{tabular}[c]{@{}c@{}}99.52\\ (1.00)\end{tabular} & \begin{tabular}[c]{@{}c@{}}0.9925 \\(0.0002)\end{tabular} & \begin{tabular}[c]{@{}c@{}}99.28\\ (0.66)\end{tabular} & \begin{tabular}[c]{@{}c@{}}0.9927 \\(0.0001)\end{tabular} & \begin{tabular}[c]{@{}c@{}}99.19\\ (2.76)\end{tabular} & \begin{tabular}[c]{@{}c@{}}0.9891 \\(0.0005)\end{tabular} & \begin{tabular}[c]{@{}c@{}}99.14 \\(4.59)\end{tabular} & \begin{tabular}[c]{@{}c@{}}0.9801 \\(0.0018)\end{tabular} \\ \hline
        \multirow{4}{*}{M$^2$LC-Net} & 100 & \begin{tabular}[c]{@{}c@{}}99.68 \\(0.07)\end{tabular} & \begin{tabular}[c]{@{}c@{}}0.8591\\ (0.0519)\end{tabular} & \begin{tabular}[c]{@{}c@{}}99.66 \\(0.05)\end{tabular} & \begin{tabular}[c]{@{}c@{}}0.8532\\ (0.0152)\end{tabular} & \begin{tabular}[c]{@{}c@{}}99.28\\ (0.07)\end{tabular} & \begin{tabular}[c]{@{}c@{}}0.7151 \\(0.0150)\end{tabular} & \begin{tabular}[c]{@{}c@{}}98.81 \\(0.18)\end{tabular} & \begin{tabular}[c]{@{}c@{}}0.6355\\ (0.0211)\end{tabular} & \begin{tabular}[c]{@{}c@{}}98.97 \\(0.40)\end{tabular} & \begin{tabular}[c]{@{}c@{}}0.6261\\ (0.0070)\end{tabular} \\
        & 50 & \begin{tabular}[c]{@{}c@{}}99.40 \\(0.04)\end{tabular} & \begin{tabular}[c]{@{}c@{}}0.8396 \\(0.0306)\end{tabular} & \begin{tabular}[c]{@{}c@{}}99.22\\ (0.08)\end{tabular} & \begin{tabular}[c]{@{}c@{}}0.7839\\ (0.0286)\end{tabular} & \begin{tabular}[c]{@{}c@{}}98.55\\ (0.06)\end{tabular} & \begin{tabular}[c]{@{}c@{}}0.6124 \\(0.0110)\end{tabular} & \begin{tabular}[c]{@{}c@{}}98.41 \\(0.09)\end{tabular} & \begin{tabular}[c]{@{}c@{}}0.5261 \\(0.0176)\end{tabular} & \begin{tabular}[c]{@{}c@{}}97.57 \\(0.11)\end{tabular} & \begin{tabular}[c]{@{}c@{}}0.4687 \\(0.0184)\end{tabular} \\
        & 10 & \begin{tabular}[c]{@{}c@{}}99.70 \\(0.10)\end{tabular} & \begin{tabular}[c]{@{}c@{}}0.9394 \\(0.0406)\end{tabular} & \begin{tabular}[c]{@{}c@{}}99.68 \\(0.04)\end{tabular} & \begin{tabular}[c]{@{}c@{}}0.9134 \\(0.0203)\end{tabular} & \begin{tabular}[c]{@{}c@{}}99.32 \\(0.04)\end{tabular} & \begin{tabular}[c]{@{}c@{}}0.7461 \\(0.0061)\end{tabular} & \begin{tabular}[c]{@{}c@{}}98.98 \\(0.05)\end{tabular} & \begin{tabular}[c]{@{}c@{}}0.6116 \\(0.0113)\end{tabular} & \begin{tabular}[c]{@{}c@{}}98.88 \\(0.08)\end{tabular} & \begin{tabular}[c]{@{}c@{}}0.5978\\ (0.0215)\end{tabular} \\
        & 1 & \begin{tabular}[c]{@{}c@{}}99.84 \\(0.03)\end{tabular} & \begin{tabular}[c]{@{}c@{}}0.9542 \\(0.0122)\end{tabular} & \begin{tabular}[c]{@{}c@{}}99.59 \\(0.01)\end{tabular} & \begin{tabular}[c]{@{}c@{}}0.8321 \\(0.0041)\end{tabular} & \begin{tabular}[c]{@{}c@{}}99.66\\ (0.01)\end{tabular} & \begin{tabular}[c]{@{}c@{}}0.8379 \\(0.0145)\end{tabular} & \begin{tabular}[c]{@{}c@{}}99.59 \\(0.02)\end{tabular} & \begin{tabular}[c]{@{}c@{}}0.8191 \\(0.0117)\end{tabular} & \begin{tabular}[c]{@{}c@{}}99.57 \\(0.03)\end{tabular} & \begin{tabular}[c]{@{}c@{}}0.8016 \\(0.0169)\end{tabular} \\ \hline
        \multirow{4}{*}{Proposed method} & 100 & \begin{tabular}[c]{@{}c@{}}99.23\\ (1.83)\end{tabular} & \begin{tabular}[c]{@{}c@{}}0.9731\\ (0.0006)\end{tabular} & \begin{tabular}[c]{@{}c@{}}98.45 \\(1.37)\end{tabular} & \begin{tabular}[c]{@{}c@{}}0.9780\\ (0.0004)\end{tabular} & \begin{tabular}[c]{@{}c@{}}97.50\\ (1.36)\end{tabular} & \begin{tabular}[c]{@{}c@{}}0.9776 \\(0.0001)\end{tabular} & \begin{tabular}[c]{@{}c@{}}97.20\\ (1.47)\end{tabular} & \begin{tabular}[c]{@{}c@{}}0.9696 \\(0.0001)\end{tabular} & \begin{tabular}[c]{@{}c@{}}96.95 \\(1.50)\end{tabular} & \begin{tabular}[c]{@{}c@{}}0.9569 \\(0.0004)\end{tabular} \\
        & 50 & \begin{tabular}[c]{@{}c@{}}99.67 \\(0.01)\end{tabular} & \begin{tabular}[c]{@{}c@{}}0.9816 \\(0.0001)\end{tabular} & \begin{tabular}[c]{@{}c@{}}99.22 \\(0.01)\end{tabular} & \begin{tabular}[c]{@{}c@{}}0.9832 \\(0.0001)\end{tabular} & \begin{tabular}[c]{@{}c@{}}98.60\\ (1.83)\end{tabular} & \begin{tabular}[c]{@{}c@{}}0.9859 \\(0.0002)\end{tabular} & \begin{tabular}[c]{@{}c@{}}97.73 \\(0.67)\end{tabular} & \begin{tabular}[c]{@{}c@{}}0.9781 \\(0.0005)\end{tabular} & \begin{tabular}[c]{@{}c@{}}97.69 \\(1.77)\end{tabular} & \begin{tabular}[c]{@{}c@{}}0.9701 \\(0.0002)\end{tabular} \\
        & 10 & \begin{tabular}[c]{@{}c@{}}99.78\\ (2.56)\end{tabular} & \begin{tabular}[c]{@{}c@{}}0.9851 \\(0.0005)\end{tabular} & \begin{tabular}[c]{@{}c@{}}99.39\\ (2.38)\end{tabular} & \begin{tabular}[c]{@{}c@{}}0.9875\\ (0.0010)\end{tabular} & \begin{tabular}[c]{@{}c@{}}98.84\\ (2.24)\end{tabular} & \begin{tabular}[c]{@{}c@{}}0.9883 \\(0.0002)\end{tabular} & \begin{tabular}[c]{@{}c@{}}98.51 \\(0.80)\end{tabular} & \begin{tabular}[c]{@{}c@{}}0.9828 \\(0.0002)\end{tabular} & \begin{tabular}[c]{@{}c@{}}98.43 \\(1.96)\end{tabular} & \begin{tabular}[c]{@{}c@{}}0.9727 \\(0.0002)\end{tabular} \\
        & 1 & \begin{tabular}[c]{@{}c@{}}99.85 \\(1.83)\end{tabular} & \begin{tabular}[c]{@{}c@{}}0.9960\\ (0.0003)\end{tabular} & \begin{tabular}[c]{@{}c@{}}99.80\\ (0.90)\end{tabular} & \begin{tabular}[c]{@{}c@{}}0.9954 \\(0.0002)\end{tabular} & \begin{tabular}[c]{@{}c@{}}99.61\\ (0.50)\end{tabular} & \begin{tabular}[c]{@{}c@{}}0.9961 \\(0.0001)\end{tabular} & \begin{tabular}[c]{@{}c@{}}99.52 \\(0.67)\end{tabular} & \begin{tabular}[c]{@{}c@{}}0.9943 \\(0.0001)\end{tabular} & \begin{tabular}[c]{@{}c@{}}99.44 \\(1.20)\end{tabular} & \begin{tabular}[c]{@{}c@{}}0.9882 \\(0.0002)\end{tabular} \\ \hline
    \end{tabular}}
\end{sc}
\end{small}
\end{center}
\vskip -0.1in
\end{table}

Table~\ref{table:accuracy_mnist_svhn} demonstrates the superior performance of the proposed method over baseline methods that jointly address multi-modal fusion and long-tailed recognition, particularly under inversely long-tailed test distributions with high imbalance ratios. For example, when the test data distribution is inversely long-tailed with an imbalance ratio of 50, the proposed method with a training imbalance ratio of 50 outperformed MMD-LA, SADE-LMF, and M$^2$LC-Net by $0.43$, $0.04$, and $0.51$, respectively, in terms of the F1 score. As the F1 score more appropriately reflects performance on imbalanced datasets than classification accuracy, these results highlight the strong generalization capability of the proposed method under severe distribution shifts.
Notably, the proposed method achieves these gains without sacrificing classification accuracy; instead, it maintains competitive accuracy compared to the baselines. In most cases, the classification accuracies of the proposed method differ from those of the baselines by within $\pm 2\%$, and in some settings, the proposed method exceeds the baselines by up to $15\%$.

The table further shows that the proposed method outperforms the original MMD and SADE methods. The original MMD does not explicitly address class imbalance in either the training or test data, while the original SADE is limited to single-modality settings. In terms of F1 score, the proposed method surpasses MMD and SADE by 0.45 and 0.19, respectively. In addition, the proposed method achieves classification accuracies comparable to MMD, with differences of approximately $\pm 1\%$, and substantially improves upon SADE, with gains of at least $7\%$. 
This improvement becomes more pronounced in challenging scenarios, reaching up to $15 \%$ when the training imbalance ratio is 50 and the test data follow an inversely long-tailed distribution with an imbalance ratio of 10. Overall, these results indicate that while MMD and SADE are each designed to address either multi-modal fusion or class imbalance in isolation, the proposed method effectively handles both simultaneously, leading to improved robustness and generalization to unseen, non-uniform data distributions.

\subsubsection{\emph{Aggregation weights for multi-expert framework}} \label{s:benchmarks_weights}
During the test phase of the proposed method, the three trained experts are combined to produce final predictions, with the aggregation weights updated to maximize prediction stability. These aggregation weights naturally favor the most suitable expert for a given test distribution. For example, under a long-tailed test data distribution, greater weight is assigned to the forward expert; for a uniform test distribution, the uniform expert receives higher weight; and for an inversely long-tailed test distribution, the backward expert is emphasized.

To illustrate how aggregation weights vary across different test data distributions, we compare the learned aggregation weights under various test settings. Since aggregation weights are specific to SADE-based methods, this comparison is conducted against two baselines: the original SADE and SADE-LMF.

Table~\ref{table:aggregation_weight_mnist_svhn} presents the aggregation weights learned by the original SADE, SADE-LMF, and the proposed method under various test data distributions. We observe that, for SADE and SADE-LMF, some aggregation weights are not assigned in accordance with the expected strengths of the corresponding experts. For example, when the test data distribution is long-tailed with an imbalance ratio of 50, the aggregation weight $w_1$ associated with the forward expert $E_1$ would be expected to exceed the weights $w_2$ and $w_3$ of the uniform expert $E_2$ and the backward expert $E_3$. However, in SADE-LMF, the aggregation weight $w_2$ for the uniform expert $E_2$ surpassed both $w_1$ and $w_3$, which is counterintuitive. 

In contrast, the proposed method assigns aggregation weights that are consistent with the underlying test data distributions. Specifically, under a long-tailed test distribution, the aggregation weight $w_1$ for the forward expert $E_1$ dominates those of the other experts. Similarly, for a uniform test distribution, the uniform expert $E_2$ receives the highest weight, while for an inversely long-tailed test distribution, the backward expert $E_3$ is assigned the largest weight.

Finally, we note that we exclude the case where the training imbalance ratio equals 1, corresponding to a balanced training dataset. In this setting, all three experts are trained to excel in handling the uniform distribution by construction, making them equally important; consequently, any expert can potentially be assigned a larger aggregation weight without a clear preference.

\begin{table}[htbp]
\caption{Aggregation weights $w_1$, $w_2$, and $w_3$ corresponding to the trained experts $E_1$, $E_2$, and $E_3$ in SADE-based methods on the \textit{MNIST} and \textit{SVHN} datasets. The abbreviation ``IR" denotes the imbalance ratio of the training data. Test data distributions range from long-tailed with an imbalance ratio 50 (Forward 50) to inversely long-tailed with an imbalance ratio 50 (Backward 50).}
\label{table:aggregation_weight_mnist_svhn}
\vskip 0.15in
\begin{center}
\begin{small}
\begin{sc}
    \resizebox{16cm}{!}{%
    \begin{tabular}{ccccccccccccccccc}
        \hline
        \multirow{2}{*}{Model} & \multirow{2}{*}{IR} & \multicolumn{3}{c}{Forward 50} & \multicolumn{3}{c}{Forward 10} & \multicolumn{3}{c}{Uniform} & \multicolumn{3}{c}{Backward 10} & \multicolumn{3}{c}{Backward 50} \\ \cline{3-17} & & $w_1$ & $w_2$ & $w_3$ & $w_1$ & $w_2$ & $w_3$ & $w_1$ & $w_2$ & $w_3$ & $w_1$ & $w_2$ & $w_3$ & $w_1$ & $w_2$ & $w_3$ \\ \hline
        \multirow{3}{*}{SADE} & 100 & 0.51 & 0.33 & 0.17 & 0.52 & 0.30 & 0.18 & 0.36 & 0.27 & 0.37 & 0.12 & 0.15 & 0.73 & 0.12 & 0.16 & 0.72 \\
        & 50 & 0.61 & 0.28 & 0.11 & 0.60 & 0.29 & 0.11 & 0.35 & 0.27 & 0.38 & 0.05 & 0.06 & 0.86 & 0.04 & 0.06 & 0.89 \\
        & 10 & 0.54 & 0.32 & 0.14 & 0.50 & 0.33 & 0.17 & 0.32 & 0.32 & 0.36 & 0.14 & 0.17 & 0.69 & 0.10 & 0.14 & 0.77 \\ \hline
        \multirow{3}{*}{SADE-LMF} & 100 & 0.31 & 0.46 & 0.23 & 0.24 & 0.47 & 0.29 & 0.05 & 0.10 & 0.86 & 0.04 & 0.05 & 0.91 & 0.04 & 0.05 & 0.91 \\
        & 50 & 0.45 & 0.37 & 0.18 & 0.53 & 0.29 & 0.18 & 0.31 & 0.36 & 0.33 & 0.18 & 0.30 & 0.52 & 0.14 & 0.31 & 0.55 \\
        & 10 & 0.37 & 0.33 & 0.30 & 0.32 & 0.34 & 0.34 & 0.24 & 0.20 & 0.56 & 0.13 & 0.14 & 0.73 & 0.16 & 0.24 & 0.60 \\ \hline
        \multirow{3}{*}{Proposed method} & 100 & 0.59 & 0.35 & 0.06 & 0.52 & 0.40 & 0.08 & 0.25 & 0.54 & 0.21 & 0.09 & 0.18 & 0.73 & 0.06 & 0.14 & 0.80 \\
        & 50 & 0.47 & 0.37 & 0.15 & 0.47 & 0.28 & 0.25 & 0.24 & 0.51 & 0.25 & 0.04 & 0.09 & 0.87 & 0.04 & 0.09 & 0.87 \\
        & 10 & 0.60 & 0.31 & 0.09 & 0.62 & 0.29 & 0.09 & 0.21 & 0.60 & 0.19 & 0.10 & 0.18 & 0.71 & 0.09 & 0.21 & 0.71 \\ \hline
    \end{tabular}}
\end{sc}
\end{small}
\end{center}
\vskip -0.1in
\end{table}

\subsection{\emph{Experiments on SIIM-ISIC Melanoma Classification Challenge dataset}} \label{s:melanoma}
\subsubsection{\emph{Datasets and experimental setup}} \label{s:melanoma_setup}
\begin{figure}[htbp]
    \begin{center}
        \begin{tabular}{c@{\hspace{0.7cm}}c}
            \includegraphics[width=2.8in]{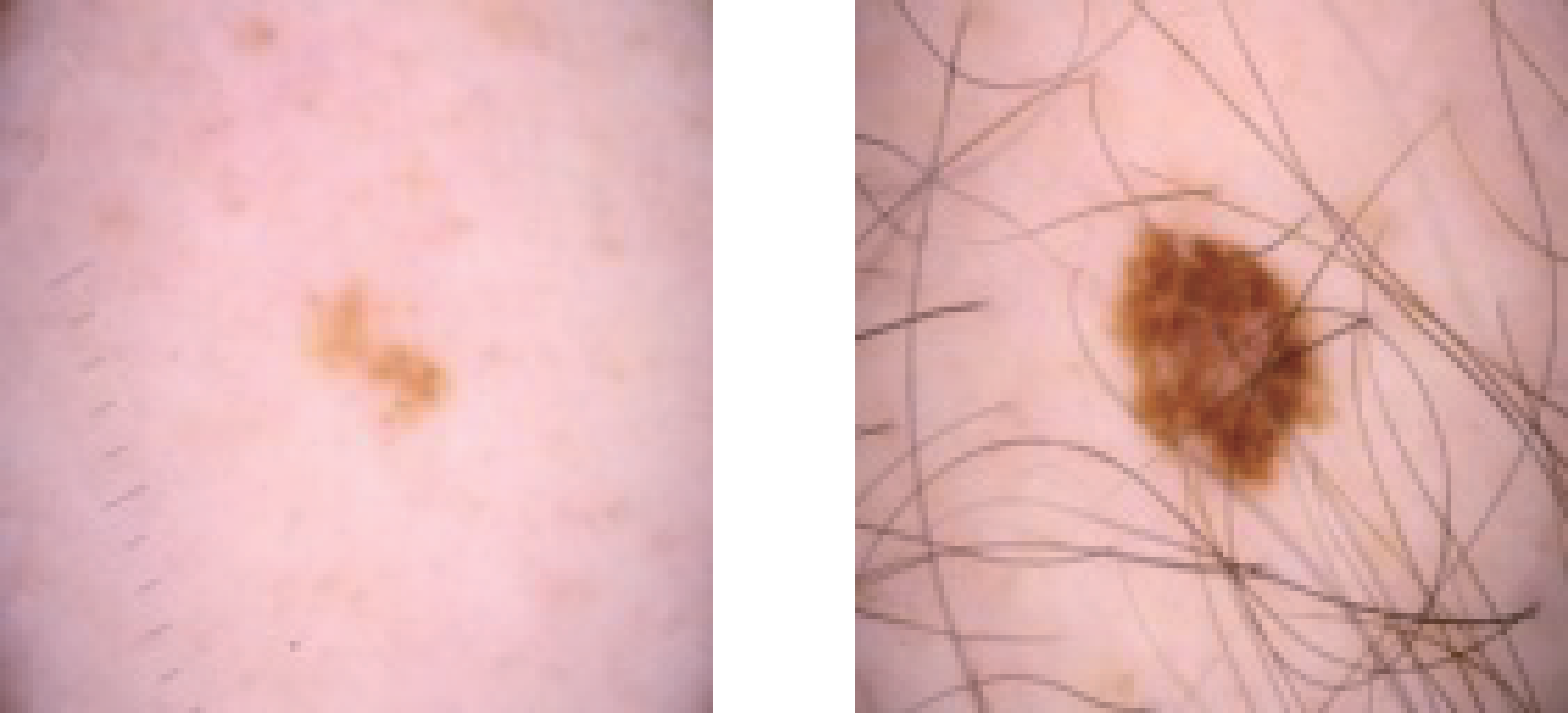} & \includegraphics[width=2.8in]{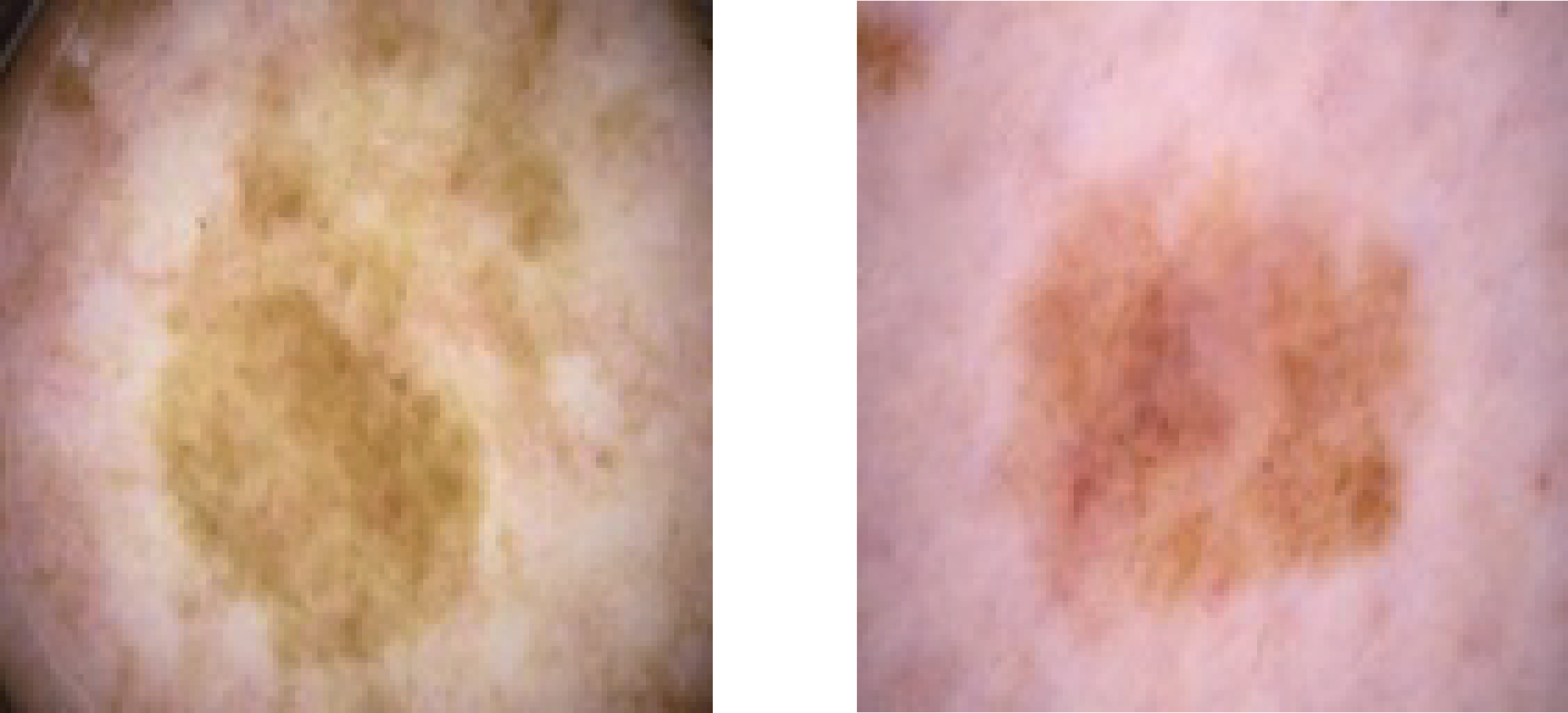}\\
            \footnotesize{(a) Benign} & \footnotesize{(b) Malignant}\\
        \end{tabular}
        \caption{Representative examples of skin lesions.} \label{fig:skin_lesions}
    \end{center}
\end{figure}

We conducted experiments on a medical dataset from the SIIM–ISIC Melanoma Classification Challenge provided by Kaggle. The dataset contains 33,126 dermoscopic images of benign and malignant skin lesions, each associated with a unique patient ID. Figure~\ref{fig:skin_lesions} shows four representative examples, where the two images on the left depict benign lesions and the two images on the right depict malignant lesions. 

The dataset also contains three additional categories of metadata for each patient: gender, age, and general anatomic site. Consequently, each sample in this dataset consists of two modalities: an image of the skin lesion and tabular metadata spanning these three categories. The dataset inherently exhibits a long-tailed distribution, comprising 32,542 benign and 584 malignant skin lesions, with an imbalance ratio of approximately 55.72.

To identify melanoma using the dataset, we first partitioned it into 26,501 training and 6,625 test data points, allocating 80$\%$ of the data points for training and the remaining 20$\%$ for testing. During this partition, we ensured that the imbalance ratio of both training and test datasets remained similar, thereby satisfying the distributional assumption for our model in the one image and one tabular modality setting. The final training dataset comprised 26,032 benign and 469 malignant skin lesions, resulting in an imbalance ratio of 55.51. Similarly, the final test dataset included 6,510 benign and 115 malignant skin lesions, with an imbalance ratio of 56.61.

For the network architecture, we employed different types of layers to extract features from image and tabular data. Specifically, image features were extracted using multiple convolutional and pooling layers followed by a series of fully connected layers. In contrast, since convolutional and pooling layers are primarily designed for image processing and are not well suited to tabular data, we replaced them with an embedding layer to extract features from the metadata.

\subsubsection{\emph{Classification results on the melanoma dataset}} \label{s:melanoma_results}
We compared the proposed model with the same baseline methods considered in Section~\ref{s:benchmarks}. As before, each experiment was repeated 10 times, and both classification accuracy and F1 score were reported along with standard errors. 
Table~\ref{table:accuracy_melanoma} summarizes the results, confirming that the proposed model consistently outperforms the baselines in terms of the F1 score. Specifically, our model achieves the highest F1 score among all methods, with improvements ranging from 0.04 to 0.08 over the baselines.
Although the proposed model does not attain the highest classification accuracy, its accuracy differs from the best-performing baseline by only $0.2\%$, which is negligible compared to the approximately $2.7\%$ gap between its accuracy and the lowest accuracy achieved by SADE.

\begin{table}[t]
\caption{Classification accuracies and F1 scores with standard errors on the melanoma dataset.}
\label{table:accuracy_melanoma}
\vskip 0.15in
\begin{center}
\begin{small}
\begin{sc}
    \begin{tabular}{ccc} \hline
    Model & Accuracy & F1 \\ \hline
    MMD & 98.28 (0.01) & 0.5250 (0.0023) \\
    SADE & 95.34 (5.46) & 0.5436 (0.0006) \\
    MMD-LA & 98.08 (0.02) & 0.5290 (0.0100) \\
    SADE-LMF & 97.87 (1.17) & 0.5629 (0.0004) \\
    M$^2$LC-Net & 98.27 (1.05) & 0.5337 (0.0150) \\
    Proposed method & 98.06 (0.01) & 0.6012 (0.0003) \\ \hline
\end{tabular}
\end{sc}
\end{small}
\end{center}
\vskip -0.1in
\end{table}

\subsubsection{\emph{Aggregation weights for multi-expert framework}} \label{s:melanoma_weights}
We analyzed the learned aggregation weights following the same procedure as in Section~\ref{s:benchmarks_weights}. Specifically, we compared the proposed model with two baseline methods: the original SADE and SADE-LMF. Table~\ref{table:aggregation_weight_melanoma} reports the aggregation weights for all three models. The results show that, across all cases, the aggregation weight $w_1$ of the forward expert $E_1$ consistently exceeds the weights $w_2$ and $w_3$ of the uniform expert $E_2$ and the backward expert $E_3$. 

Moreover, when comparing the aggregation weight $w_1$ across models,we observe that the proposed model assigns a larger weight to the forward expert than SADE-LMF, which in turn assigns a larger weight than SADE. Conversely, the aggregation weight $w_3$ of the proposed model is lower than that of SADE-LMF, which is lower than that of SADE. 
These trends indicate that the experts in the proposed model are more effectively trained than those in SADE and SADE-LMF. In particular, the forward expert exhibits enhanced capability in handling long-tailed data distributions, while each expert more strongly specializes in the distribution it is designed to model.

\begin{table}[htbp]
\caption{Aggregation weights $w_1$, $w_2$, and $w_3$ associated with the trained experts $E_1$, $E_2$, and $E_3$ in the SADE-based methods on the melanoma dataset.}
\label{table:aggregation_weight_melanoma}
\vskip 0.15in
\begin{center}
\begin{small}
\begin{sc}
    \begin{tabular}{cccc} \hline
    Model & $w_1$ & $w_2$ & $w_3$ \\ \hline
    SADE & 0.34 & 0.34 & 0.31 \\
    SADE-LMF & 0.44 & 0.32 & 0.24 \\
    Proposed & 0.66 & 0.21 & 0.13 \\ \hline
\end{tabular}
\end{sc}
\end{small}
\end{center}
\vskip -0.1in
\end{table}

\section{Conclusion}\label{s:conclusion}
In this paper, we proposed a new method for long-tailed recognition with multi-modal data. The proposed method adopts a multi-expert framework and modifies the expert architecture to jointly integrate all input modalities within each expert. A key component of the proposed method is a classifier module that quantifies the relative importance of each modality during multi-modal fusion. In addition, we introduced a tailored training and testing strategy for tabular data, in which the aggregation weights are learned during training, thereby eliminating the need for self-supervised optimization at test time.
We validated the effectiveness of the proposed method through extensive experiments on both benchmark image datasets and a real-world dataset comprising image and tabular modalities. The experimental results demonstrate that the proposed method is well suited for classifying samples from multi-modal datasets with long-tailed distributions.
In future work, we plan to extend our framework to partially labeled settings by incorporating semi-supervised learning strategies. Addressing this scenario is expected to further alleviate class imbalance in practical applications, where labeled data are often limited and unevenly distributed.

\section{Data Availability Statement}
The benchmark image datasets used in this study are available at \\https://yann.lecun.com/exdb/mnist and https://ufdl.standford.edu/housenumbers. The datasets are also accessible through the Pytorch library in Python. The SIIM-ISIC Melanoma Classification Challenge dataset is available at https://www.kaggle.com/c/siim-isic-melanoma-classification/data.

\clearpage
\newpage

\bibliographystyle{abbrv}
\bibliography{reference}

\end{document}